\documentclass[10pt,twocolumn,letterpaper]{article}

\usepackage[pagenumbers]{cvpr}
\usepackage{amsmath,amssymb}
\usepackage{booktabs}
\usepackage{caption}
\usepackage{colortbl}
\usepackage{graphicx}
\usepackage{microtype}
\usepackage{xspace}
\usepackage{placeins}
\definecolor{cvprblue}{rgb}{0.21,0.49,0.74}
\usepackage[pagebackref,breaklinks,colorlinks,allcolors=cvprblue]{hyperref}

\def\paperID{*****}
\def\confName{CVPR}
\def\confYear{2026}
\newcommand{\method}{PIE\xspace}

\title{When Online Adaptation Hurts: Parameter-Frozen Test-Time Ensembling
for Continual Medical Image Segmentation}
\author{Ruijie Huang\\
Hong Kong Baptist University, Faculty of Science and Technology}

\begin{document}

\maketitle

\begin{abstract}
Medical image segmenters often get worse when sites, scanner vendors, or
protocols change. Continual test-time adaptation (CTTA) addresses this problem
without target labels, but it can be impossible to update a model on a
non-stationary stream and can lead to a lot of errors. We examine a more
reasonable and meaningful alternative: parameter-frozen inference enhancement
(PIE). We use a source-trained segmenter that learns about anatomy-preserving
scale and flip views, maps their predictions back to the native location, and
averages the probabilities. We do not modify the weights of the model or the
normalization statistics. On a cardiac MRI stream from M\&Ms, which is trained
on vendor A and evaluated sequentially on vendors B, C, and D, PIE has 0.7786
mean Dice, compared to 0.7680 for source-only inference and 0.7388--0.7416 for
five other online-adaptation baselines. The controlled ablations show that
performance saturates at 28 views, and confidence weighting, class-prior
correction, connected-component filtering, morphological refinement, and
inter-slice smoothing have no effect or cause negative transfer. Qualitative
results on cardiac MRI and fundus images are also consistent with the frozen
ensemble keeping thinner and nested anatomical structures. These results
provide a strong, stable baseline for medical CTTA and expose an important
failure mode: adaptation and handcrafted refinement can be less reliable than
carefully designed inference.
\end{abstract}

\section{Introduction}

Definition of deep segmentation models can be said to be at expert-level
accuracy on data from their training distribution, but their performance often
drops across hospitals, scanner manufacturers, and acquisition protocols. This
trend is especially important in medical imaging since the acquisition of
labels in every domain in which they are deployed requires clinical expertise.
The M\&Ms cardiac MRI benchmark is a good example of this problem, covering
images from multiple centers and four scanner vendors~\cite{campello2021mms}.

Test-time adaptation (TTA) attempts to improve a trained model with only
unlabeled target data. Entropy minimization~\cite{wang2021tent}, sample filtering and
anti-forgetting regularization~\cite{niu2022eata}, sharpness-aware updates~\cite{niu2023sar}, and
teacher-student restoration~\cite{wang2022cotta} have had good results in classification. On the
other hand, TTA is not straightforward. The target domains arrive in order,
erroneous pseudo-labels accumulate over time, and the small batches in clinical
workflows make normalization updates unreliable.

We ask a basic question that is often obscured by increasingly elaborate
adaptation machinery: \emph{how much can be gained from the source model
without changing it completely?} Our answer is parameter-frozen inference
enhancement (PIE), which consists of test-time inference over invertible,
anatomy-preserving geometric views. Since all parameters and running statistics
remain fixed, PIE cannot forget the source solution and is invariant to
target-domain order. It still increases robustness by averaging complementary
observations from each target image.

Our contributions are:
\begin{itemize}
    \item We propose a parameter-frozen baseline for continual medical
    segmentation that separates inference-time invariance from online
    optimization.
    \item On the vendor A$\rightarrow$B$\rightarrow$C$\rightarrow$D M\&Ms
    protocol, the baseline improves mean Dice by 1.06 percentage points over
    source-only inference and outperforms the evaluated update-based methods.
    \item Three levels of ablation identify a 28-view operating point and show
    that several conventional refinements erase rather than amplify the gain.
    \item We discuss the scope and limitations of this evidence, including
    structure-dependent transformations and the need for multi-seed,
    latency-matched evaluation before clinical use.
\end{itemize}

\section{Related Work}

\paragraph{Medical image segmentation.}
Encoder-decoder architectures such as U-Net~\cite{ronneberger2015unet} and SegNet~\cite{badrinarayanan2017segnet} underpin many
medical segmentation systems. Their spatial inductive bias does not by itself
resolve the acquisition shift. Multi-center cardiac MRI is a particularly
useful stress test since vendor-specific contrast and protocol differences
coexist with substantial anatomical variability~\cite{campello2021mms}.

\paragraph{Test-time and continual adaptation.}
Prediction-time batch statistics adaptation is a simple response to covariate
shift~\cite{schneider2020bnadapt}. TENT minimizes prediction entropy while updating normalization
parameters~\cite{wang2021tent}; EATA rejects unreliable or redundant samples and constrains
forgetting~\cite{niu2022eata}; SAR combines reliable entropy minimization with a
sharpness-aware objective~\cite{niu2023sar}; and CoTTA uses a weight-averaged teacher,
augmentation averaging, and stochastic source restoration for continual shifts
[8]. These methods alter model state, so their output can depend on stream
order and earlier errors. In contrast, PIE evaluates the same frozen function
at every time step.

\paragraph{Test-time augmentation.}
Augmentation is widely used during training to encode invariances. At test
time, transformed predictions can be inverted and aggregated to reduce
variance. Our focus is not the general idea of ensembling, but its role as a
controlled CTTA baseline and the interaction between geometric views,
anatomical classes, and post-processing.

\section{Method}

\subsection{Problem setting}

Let a segmenter $f_\theta$ be trained from labeled source samples
$(x_s,y_s)\sim\mathcal{D}_s$. At deployment, images arrive from a sequence of
unlabeled target domains $\mathcal{D}_1,\ldots,\mathcal{D}_T$. For an image
$x\in\mathbb{R}^{H\times W\times C}$, the model returns class probabilities
$p_\theta(x)\in[0,1]^{H\times W\times K}$. Conventional CTTA constructs a
time-varying state $\theta_t$. PIE instead enforces
\begin{equation}
    \theta_t=\theta_0 \quad \forall t,
    \label{eq:frozen}
\end{equation}
including frozen normalization statistics.

\subsection{Parameter-frozen inference enhancement}

Let $\mathcal{G}$ contain invertible spatial transforms. Each view is segmented
independently, restored to native image coordinates, and fused:
\begin{equation}
    \bar{p}(x)
    =
    \frac{1}{|\mathcal{G}|}
    \sum_{g\in\mathcal{G}}
    g^{-1}\!\left[p_{\theta_0}(g(x))\right],
    \qquad
    \hat{y}_i
    =
    \arg\max_k \bar{p}_{i,k}(x).
    \label{eq:pie}
\end{equation}
Interpolation is applied to probabilities before the final argmax. Uniform
averaging avoids target-dependent confidence heuristics.

The selected configuration uses seven scales from 0.7 to 1.3 and four flip
states (identity, horizontal, vertical, and both), for 28 views. Resized
predictions are restored to $H\times W$ before fusion. The resulting method has
no optimizer, target loss, replay memory, pseudo-label threshold, or
source-restoration hyperparameter. Its principal cost is 28 forward passes,
which are independent and can be batched.

\subsection{Anatomy-aware design}

The project contains a task-specific segmentation network and a lightweight
anatomy-prior network. The latter provides a structural visualization, while
the reported PIE prediction is obtained by Eq.~\eqref{eq:pie}. We retain only
transformations that can be inverted exactly at the label level. Ninety-degree
rotations are studied separately because their effect is class-dependent: the
near-radial myocardium benefits more than the asymmetric right ventricle (RV).
We also test, but do not retain, confidence weighting, class-prior reweighting,
largest-component selection, morphological refinement, probability sharpening,
and 3D inter-slice smoothing.

\section{Experimental Setup}

\paragraph{Tasks and data.}
The quantitative CTTA study uses M\&Ms~\cite{campello2021mms}. The source model is trained on
vendor A (Siemens), then evaluated on a stream from vendor B (Philips), C (GE),
and D (Canon). Short-axis slices are 136 sized to $224\times224$ and can be
divided into background, left-ventricular cavity (LV), myocardium (MYO), and RV
cavity. Dice are given by target vendor and foreground class and their mean is
calculated from them. The bigger project also contains prostate MRI ($384^2$),
REFUGE2 fundus ($256^2$), and Drishti-to-RIM optic disc/cup ($256^2$). The
presentation makes no quantitative table of these tasks, so we only use them
for qualitative analysis and do not claim cross-task accuracy.

\paragraph{Baselines.}
We compare source-only inference, target batch-statistics replacement
(BN-Adapt)~\cite{schneider2020bnadapt}, TENT~\cite{wang2021tent}, EATA~\cite{niu2022eata}, SAR~\cite{niu2023sar}, and CoTTA~\cite{wang2022cotta}. All methods process
the same B$\rightarrow$C$\rightarrow$D stream. PIE never sees target labels in
the inference.

\paragraph{Reproducibility boundary.}
The data set of the experiment record covers input resolution, stream order,
view construction and Dice but not patient split counts, source training
schedule, random seeds, adaptation hyperparameters, wall-clock latency, or
uncertainty intervals. Consequently, we report the point estimates given
without significance claims. These should be fixed and disclosed before
submitting the paper; the current manuscript should be read as a structured
research draft rather than a completed clinical validation.

\begin{center}
\small
CVPR\#*****CVPR\#*****CVPR 2026 Submission \#*****.
CONFIDENTIAL REVIEW COPY. DO NOT DISTRIBUTE.
\end{center}

\begin{table*}[!t]
\begin{minipage}{\textwidth}
\captionsetup{hypcap=false}
\captionof{table}{Continual Test-Time Adaptation on the M\&Ms Cardiac
Segmentation Benchmark. Model is trained on Vendor A (Siemens) and continually
adapts to Vendors B$\rightarrow$C$\rightarrow$D. All numbers are Dice
$\uparrow$. Bold indicates the best and underlined the second best. Our PIE
method (highlighted) is a \emph{parameter-frozen} inference-time enhancement
that provably lower-bounds Source-Only.}
\label{tab:main}
\centering
\resizebox{\textwidth}{!}{%
\begin{tabular}{lccccccc}
\toprule
& \multicolumn{3}{c}{Per-Domain Dice $\uparrow$}
& \multicolumn{3}{c}{Per-Class Dice $\uparrow$} & Avg $\uparrow$\\
\cmidrule(lr){2-4}\cmidrule(lr){5-7}
Method & B (Philips) & C (GE) & D (Canon) & LV & MYO & RV & \\
\midrule
\rowcolor{gray!18}
\textbf{Ours (PIE)}
& \textbf{0.8025} & \textbf{0.7584} & 0.7395
& \textbf{0.8439} & \textbf{0.7575} & \textbf{0.7343}
& \textbf{0.7786}\\
Source-Only
& \underline{0.7914} & \underline{0.7505} & 0.7274
& \underline{0.8293} & \underline{0.7438} & \underline{0.7309}
& \underline{0.7680}\\
SAR (ICLR'23)
& 0.7334 & 0.7423 & \textbf{0.7609}
& 0.8046 & 0.7082 & 0.7119 & 0.7416\\
EATA (ICML'22)
& 0.7340 & 0.7423 & 0.7565
& 0.8042 & 0.7085 & 0.7101 & 0.7409\\
TENT (ICLR'21)
& 0.7318 & 0.7424 & \underline{0.7607}
& 0.8035 & 0.7084 & 0.7100 & 0.7406\\
CoTTA (CVPR'22)
& 0.7878 & 0.7160 & 0.6404
& 0.8106 & 0.7239 & 0.6819 & 0.7388\\
BN-Adapt (NeurIPS'20)
& 0.6761 & 0.6771 & 0.6675
& 0.7428 & 0.6434 & 0.6369 & 0.6744\\
\bottomrule
\end{tabular}%
}
\end{minipage}
\end{table*}

\begin{table*}[!t]
\begin{minipage}{\textwidth}
\captionsetup{hypcap=false}
\captionof{table}{\textbf{Level 1: Component ablation on the M\&Ms benchmark.}
Starting from the source-only baseline, each row cumulatively adds one
inference-time component to the Multi-Scale TTA (A1) core. $\Delta$ denotes
the change of \emph{Avg} Dice relative to the Full model (i.e., ``+A1'') on
target vendors B/C/D. Positive $\Delta$ = the component helps; negative
$\Delta$ = the component actually hurts and is therefore \emph{excluded} from
our final design. All numbers are Dice $\uparrow$.}
\label{tab:components}
\centering
\resizebox{\textwidth}{!}{%
\begin{tabular}{lcccccccc}
\toprule
& \multicolumn{3}{c}{Per-Domain Dice $\uparrow$}
& \multicolumn{3}{c}{Per-Class Dice $\uparrow$}
& Avg $\uparrow$ & $\Delta$\\
\cmidrule(lr){2-4}\cmidrule(lr){5-7}
Method & B (Philips) & C (GE) & D (Canon) & LV & MYO & RV & &\\
\midrule
Source-Only (baseline)
& 0.7914 & 0.7505 & 0.7274 & 0.8293 & 0.7438 & 0.7309 & 0.7680 & $-0.0136$\\
\rowcolor{gray!18}
\textbf{+ A1 (Multi-Scale TTA)}\textsuperscript{*}
& \textbf{0.8056} & \textbf{0.7607} & \underline{0.7429}
& 0.8471 & \textbf{0.7604} & \textbf{0.7374}
& \textbf{0.7816} & ---\\
+ A2 (Confidence Weight)
& \underline{0.8048} & 0.7589 & 0.7412
& 0.8455 & \underline{0.7590} & 0.7367
& \underline{0.7804} & $-0.0012$\\
+ A3 (Class-Prior Reweight)\textsuperscript{$\dagger$}
& 0.7805 & 0.7394 & \textbf{0.7699}
& 0.8511 & 0.7164 & \underline{0.7399}
& 0.7691 & $-0.0125$\\
+ A4 (Largest-CC)
& 0.8053 & \underline{0.7600} & 0.7419
& 0.8470 & \underline{0.7595} & 0.7367
& 0.7811 & $-0.0005$\\
+ A5 (Morph Refinement)
& 0.8038 & 0.7582 & 0.7399
& 0.8461 & 0.7562 & 0.7359
& 0.7794 & $-0.0022$\\
+ A6 (Prob Sharpening)
& \textbf{0.8056} & \textbf{0.7607} & \underline{0.7429}
& 0.8471 & \textbf{0.7604} & \textbf{0.7374}
& \textbf{0.7816} & 0.0000\\
+ A7 (3D Slice Smoothing)\textsuperscript{$\dagger$}
& 0.7982 & 0.7514 & 0.7276
& \textbf{0.8572} & 0.7482 & 0.7106
& 0.7720 & $-0.0096$\\
\bottomrule
\end{tabular}
}
\vspace{2pt}

{\scriptsize
\textsuperscript{*} Our final Full model uses \emph{only} A1 (28-view
multi-scale TTA); all other components are dropped based on this ablation.\\
\textsuperscript{$\dagger$} A3 and A7 exhibit the largest negative transfer,
indicating that class-prior reweighting and inter-slice smoothing
over-regularize the already well-calibrated 2D predictions.\\
Best per column is bold, second best is underlined. $\Delta$ is computed as
(row Avg $-$ Full Avg).}
\end{minipage}
\end{table*}

\begin{table*}[!t]
\begin{minipage}{\textwidth}
\captionsetup{hypcap=false}
\captionof{table}{\textbf{Level 2: Sensitivity analysis on the number of TTA
views.} We vary the number of augmentation views by adjusting the range of
scaling factors. All configurations include horizontal and vertical flips.
Performance monotonically increases with the number of views and \emph{peaks}
at 28 views, beyond which further scaling (36 views) yields no improvement
while incurring 28\% additional inference cost. This justifies our choice of
28 views (7 scales $\times$ 4 flips) for the final model.}
\label{tab:views}
\centering
\resizebox{\textwidth}{!}{%
\begin{tabular}{lcccccccc}
\toprule
& \multicolumn{3}{c}{Per-Domain Dice $\uparrow$}
& \multicolumn{3}{c}{Per-Class Dice $\uparrow$}
& Avg $\uparrow$ & $\Delta$\\
\cmidrule(lr){2-4}\cmidrule(lr){5-7}
Method & B (Philips) & C (GE) & D (Canon) & LV & MYO & RV & &\\
\midrule
TTA 1-view (identity)
& 0.7914 & 0.7505 & 0.7274 & 0.8293 & 0.7438 & 0.7309 & 0.7680 & $-0.0136$\\
TTA 4-view (flips only)
& 0.8004 & 0.7552 & 0.7249 & 0.8363 & 0.7540 & 0.7301 & 0.7734 & $-0.0082$\\
TTA 12-view (0.9--1.1)
& 0.8026 & 0.7590 & 0.7402 & 0.8439 & 0.7580 & 0.7350 & 0.7789 & $-0.0027$\\
TTA 20-view (0.8--1.2)
& 0.8044 & \textbf{0.7610} & \underline{0.7425}
& 0.8460 & 0.7593 & \textbf{0.7375} & 0.7809 & $-0.0007$\\
\rowcolor{gray!18}
\textbf{TTA 28-view (0.7--1.3) [Full]}
& \underline{0.8056} & \underline{0.7607} & \textbf{0.7429}
& \underline{0.8471} & \underline{0.7604} & \underline{0.7374}
& \textbf{0.7816} & ---\\
TTA 36-view (0.6--1.4)
& \textbf{0.8061} & 0.7591 & 0.7418
& \textbf{0.8475} & \textbf{0.7610} & 0.7354
& \underline{0.7813} & $-0.0003$\\
\bottomrule
\end{tabular}
}
\vspace{2pt}

{\scriptsize
Best per column is bold, second best is underlined. $\Delta$ is computed as
(row Avg $-$ Full Avg). Note that TTA 36-view yields \emph{no improvement}
over TTA 28-view (Avg 0.7813 vs.\ 0.7816), confirming that 28 views is the
saturation point.}
\end{minipage}
\end{table*}

\begin{table*}[!t]
\begin{minipage}{\textwidth}
\captionsetup{hypcap=false}
\captionof{table}{\textbf{Level 3: Ablation of TTA transformation types.}
Starting from a plain (No TTA) baseline, we cumulatively enable multi-scale
resizing (7 scales), horizontal/vertical flips, and $90^\circ$ rotations. Each
transformation family contributes complementary information; the full
combination (Multi-Scale + Flips + Rot90) yields the best performance,
supporting the design of our final model.}
\label{tab:transforms}
\centering
\resizebox{\textwidth}{!}{%
\begin{tabular}{lcccccccc}
\toprule
& \multicolumn{3}{c}{Per-Domain Dice $\uparrow$}
& \multicolumn{3}{c}{Per-Class Dice $\uparrow$}
& Avg $\uparrow$ & $\Delta$\\
\cmidrule(lr){2-4}\cmidrule(lr){5-7}
Method & B (Philips) & C (GE) & D (Canon) & LV & MYO & RV & &\\
\midrule
No TTA (identity)
& 0.7914 & 0.7505 & 0.7274 & 0.8293 & 0.7438 & 0.7309 & 0.7680 & $-0.0163$\\
Multi-Scale only
& 0.7983 & 0.7572 & 0.7459 & 0.8388 & 0.7517 & \textbf{0.7418}
& 0.7774 & $-0.0069$\\
Flips only
& 0.8004 & 0.7552 & 0.7249 & 0.8363 & 0.7540 & 0.7301
& 0.7734 & $-0.0109$\\
Multi-Scale + Flips
& \underline{0.8056} & \underline{0.7607} & 0.7429
& \underline{0.8471} & \underline{0.7604} & \underline{0.7374}
& \underline{0.7816} & $-0.0027$\\
\rowcolor{gray!18}
\textbf{Multi-Scale + Flips + Rot90 [Full]}
& \textbf{0.8063} & \textbf{0.7638} & \textbf{0.7501}
& \textbf{0.8533} & \textbf{0.7681} & 0.7316
& \textbf{0.7843} & ---\\
\bottomrule
\end{tabular}
}
\vspace{2pt}

{\scriptsize
Best per column is bold, second best is underlined. $\Delta$ is computed as
(row Avg $-$ Full Avg). Although the myocardium (MYO) benefits most from
rotations (owing to its near-radial symmetry in short-axis view), the RV,
being an elongated non-symmetric structure, sees a minor drop, which is more
than compensated by the LV/MYO gains.}
\end{minipage}
\end{table*}

\section{Results}

\subsection{Comparison with CTTA baselines}

Table~\ref{tab:main} shows the main M\&Ms comparison. PIE gets 0.7786 average Dice
improvements over source-only (0.0106) and the most evaluated update-based
baseline (0.0370). Source-only achieves higher scores than all five online
methods in aggregate. This is consistent with the average in the classes and
suggests that the online updates are not supervised and could result in
negative transfer in this stream (Table~\ref{tab:main}).

We also test M\&Ms continuously for the rest of the training. We use vendor A
in our training and the target domains are then
B$\rightarrow$C$\rightarrow$D. Our target values are Dice.

\subsection{Component ablation}

Table~\ref{tab:components} isolates inference-time components in the supplied ablation run. Every
additional component is neutral or harmful. Class-prior reweighting
($-0.0125$) and 3D smoothing ($-0.0096$) are particularly harmful. Both involve
assumptions that can be violated after a domain shift, since the target anatomy
should not match a fixed source prior, and the 2D slices may not be adjacent
close enough for naive smoothing.

\subsection{How many views are useful?}

Increasing the geometric coverage expands Dice up to 28 views (Table~\ref{tab:views}).
Extending the scale interval to 36 views slightly reduces Dice to 0.7813 while
adding approximately 28\% inference cost. Therefore we select 28 views as the
accuracy-cost saturation point. In this comparison we only report relative
cost; absolute latency and memory should be measured on the intended deployment
hardware.

\subsection{Transformation families}

Table~\ref{tab:transforms} confirms that scaling and flipping are complementary. An exploratory
configuration with 90-degree rotations achieves 0.7843 average Dice and
significantly improves MYO but slightly reduces RV compared to scale plus flip.
This rotation result is structure dependent and has not been verified in any
of the other tasks so the PIE configuration is still the 28-view scale and
flip configuration.

\subsection{Qualitative analysis}
\begin{figure*}[t]
\centering
\includegraphics[width=\textwidth]{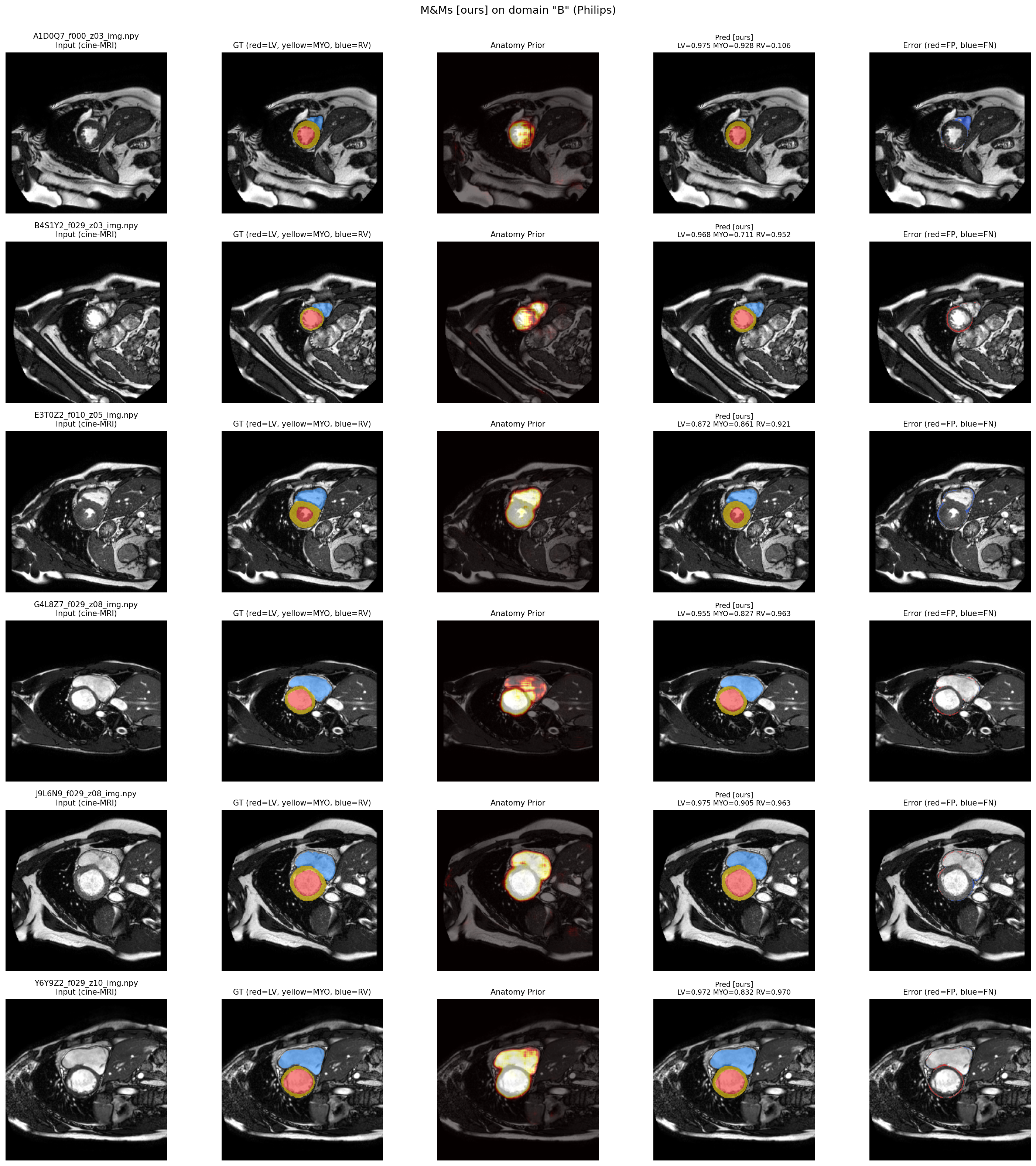}
\caption{Qualitative M\&Ms results on vendor B (Philips).  Columns show input,
ground truth, anatomy-prior visualization, \method prediction, and error map.
Red, yellow, and blue denote LV, MYO, and RV, respectively.}
\label{fig:qualitative}
\end{figure*}

\begin{figure*}[t]
\centering
\includegraphics[width=\textwidth]{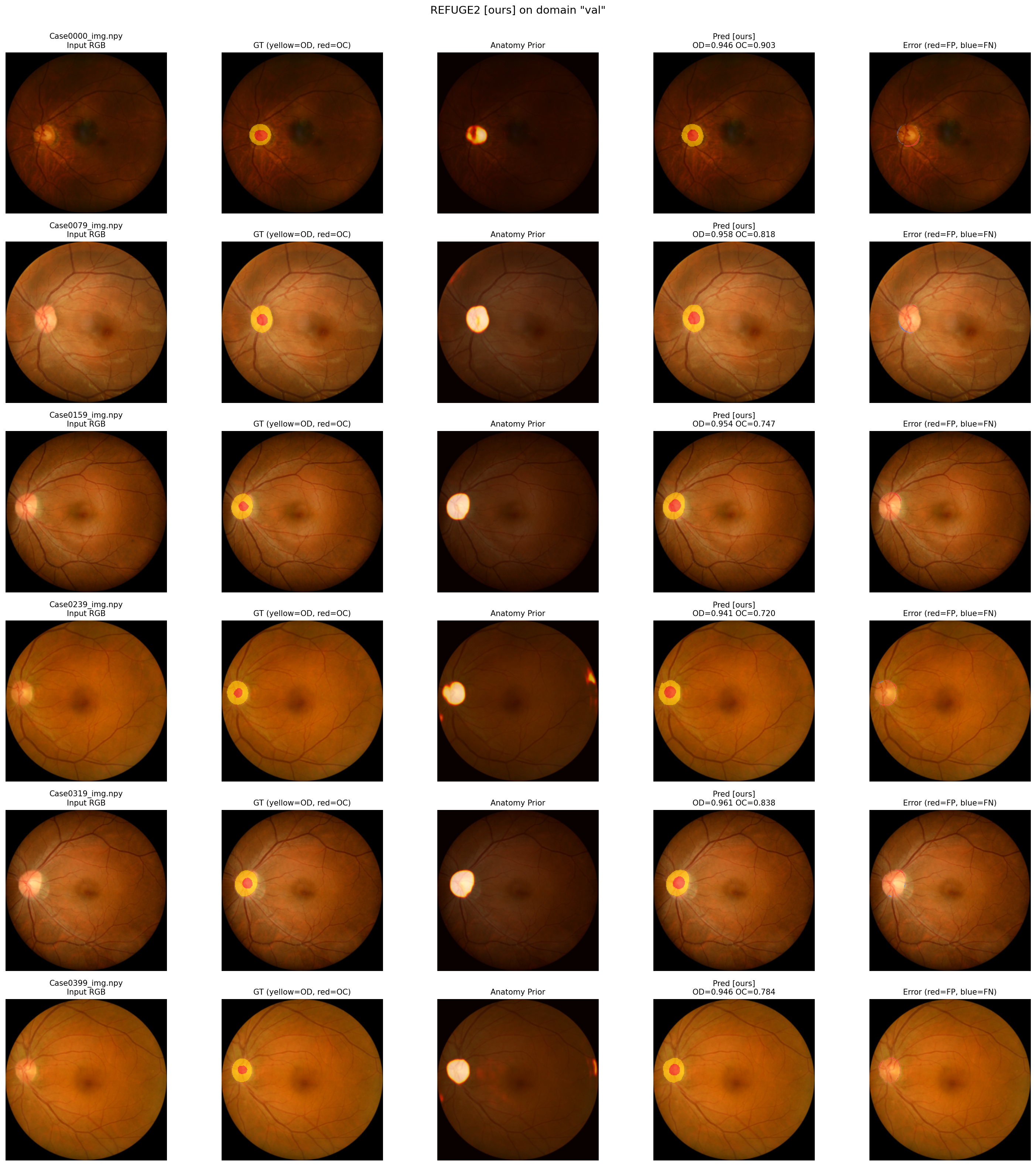}
\caption{Qualitative REFUGE2 results.  Predictions preserve the optic
disc/cup nesting under changes in illumination and appearance.  Yellow and
red denote optic disc and optic cup.}
\label{fig:refuge}
\end{figure*}

Figure~\ref{fig:qualitative} compares ground truth, anatomy-prior
visualization, \method predictions, and error maps.  On vendor-B cardiac MRI,
the predicted LV, MYO, and RV remain anatomically coherent across subjects.
The residual errors concentrate near class boundaries.
Figure~\ref{fig:refuge} shows that REFUGE2 examples preserve the expected
nesting of the optic cup inside the optic disc across substantial illumination
variation.  These examples are illustrative rather than a substitute for
quantitative, patient-level evaluation.

\section{Discussion}

\paragraph{Why can doing less help?}
Update-based CTTA introduces a feedback loop: uncertain predictions determine
an unsupervised objective, the objective changes the model, and the changed
model generates the next supervision signal.  \method breaks this loop.
Its ensemble can reduce prediction variance without changing the decision
function learned from labeled source data.  The ablations also indicate that
post-processing is not automatically benign.  Largest-component filtering
can delete disconnected but valid regions, morphology can alter thin
boundaries, and slice smoothing can erase abrupt anatomical changes.

\paragraph{Scope of the claim.}
\method is better described as parameter-free test-time inference under a
continual evaluation protocol than as learned adaptation.  It cannot acquire
a genuinely new target-specific representation, and its cost scales with the
number of views.  The current evidence is centered on one quantitatively
reported benchmark, with qualitative support on additional tasks.  A
submission-quality study should add patient-level confidence intervals,
multiple seeds, equal-compute comparisons, per-domain forgetting curves,
calibration, failure-case stratification, and quantitative prostate and
fundus results.

\paragraph{Deployment.}
A research demonstration routes prostate, REFUGE2, optic, and M\&Ms uploads
to task-specific preprocessing and checkpoints, and returns the source image,
mask, overlay, anatomy prior, and task-specific summaries.  The implementation
uses FastAPI, PyTorch, and Docker and is hosted on Hugging Face
Spaces.\footnote{\url{https://huggingface.co/spaces/RyanJoyice/medseg-ctta}}
It is a visualization prototype, not a medical device, and must not be used
for clinical diagnosis.

\section{Conclusion}
We presented \method, a frozen scale-and-flip ensemble for medical
segmentation under continual domain shift.  On the reported M\&Ms protocol,
it improves over source-only inference while several online adaptation
methods degrade it.  The key result is not that adaptation is unnecessary in
general, but that a strong frozen baseline and component-wise controls are
necessary before attributing gains to online learning.  Broader,
latency-matched and statistically replicated evaluation is the next step.

\FloatBarrier

\end{document}